\documentclass[10pt,twocolumn,letterpaper]{article}

\usepackage[pagenumbers]{iccv} %

\definecolor{iccvblue}{rgb}{0.21,0.49,0.74}
\usepackage[pagebackref,breaklinks,colorlinks,allcolors=iccvblue]{hyperref}
\usepackage{float}
\usepackage{caption}
\usepackage{lipsum}
\usepackage{booktabs}
\usepackage{siunitx}
\usepackage{caption}
\usepackage{xcolor}
\usepackage{multirow}
\usepackage{booktabs}
\usepackage{dsfont}
\usepackage{placeins} %
\usepackage{afterpage}
\usepackage{graphicx}  %
\usepackage{amssymb}   %
\usepackage{colortbl}
\usepackage{pifont}
\usepackage{tabularx}
\usepackage{booktabs}   %
\usepackage{comment}
\usepackage{fontawesome5}
\def\paperID{21} %
\def\confName{ICCV}
\def\confYear{2025}

\title{Towards Effective Human-in-the-Loop Assistive AI Agents}

\author{
Filippos Bellos$^{1,}$\textsuperscript{\faGlasses} \quad
Yayuan Li$^{1,}$\textsuperscript{\faGlasses} \quad
Cary Shu$^1$ \quad
Ruey Day$^1$\\
Jeffrey M. Siskind$^2$ \quad
Jason J. Corso$^{1,3}$\\
\\
$^1$University of Michigan \quad
$^2$Purdue University \quad
$^3$Voxel51\\
\textsuperscript{\faGlasses}Equal Contribution\\
{\tt\small \{fbellos, yayuanli\}@umich.edu}
}

\begin{document}

\maketitle

\begin{abstract}

    Effective human-AI collaboration for physical task completion has significant potential in both everyday activities and professional domains. AI agents equipped with informative guidance can enhance human performance, but evaluating such collaboration remains challenging due to the complexity of human-in-the-loop interactions. In this work, we introduce an evaluation framework and a multimodal dataset of human-AI interactions designed to assess how AI guidance affects procedural task performance, error reduction and learning outcomes.
    Besides, we develop an augmented reality (AR)-equipped AI agent that provides interactive guidance in real-world tasks, from cooking to battlefield medicine. Through human studies\footnote{The Institutional Review Board (IRB) of our institution approved this human subjects research before the start of the study.}, we share empirical insights into AI-assisted human performance and demonstrate that AI-assisted collaboration improves task completion.
\end{abstract}

\section{Introduction}

\begin{figure}[t] 
    \centering
    \hspace*{-0.7cm} %
    \includegraphics[width=0.4\textwidth]{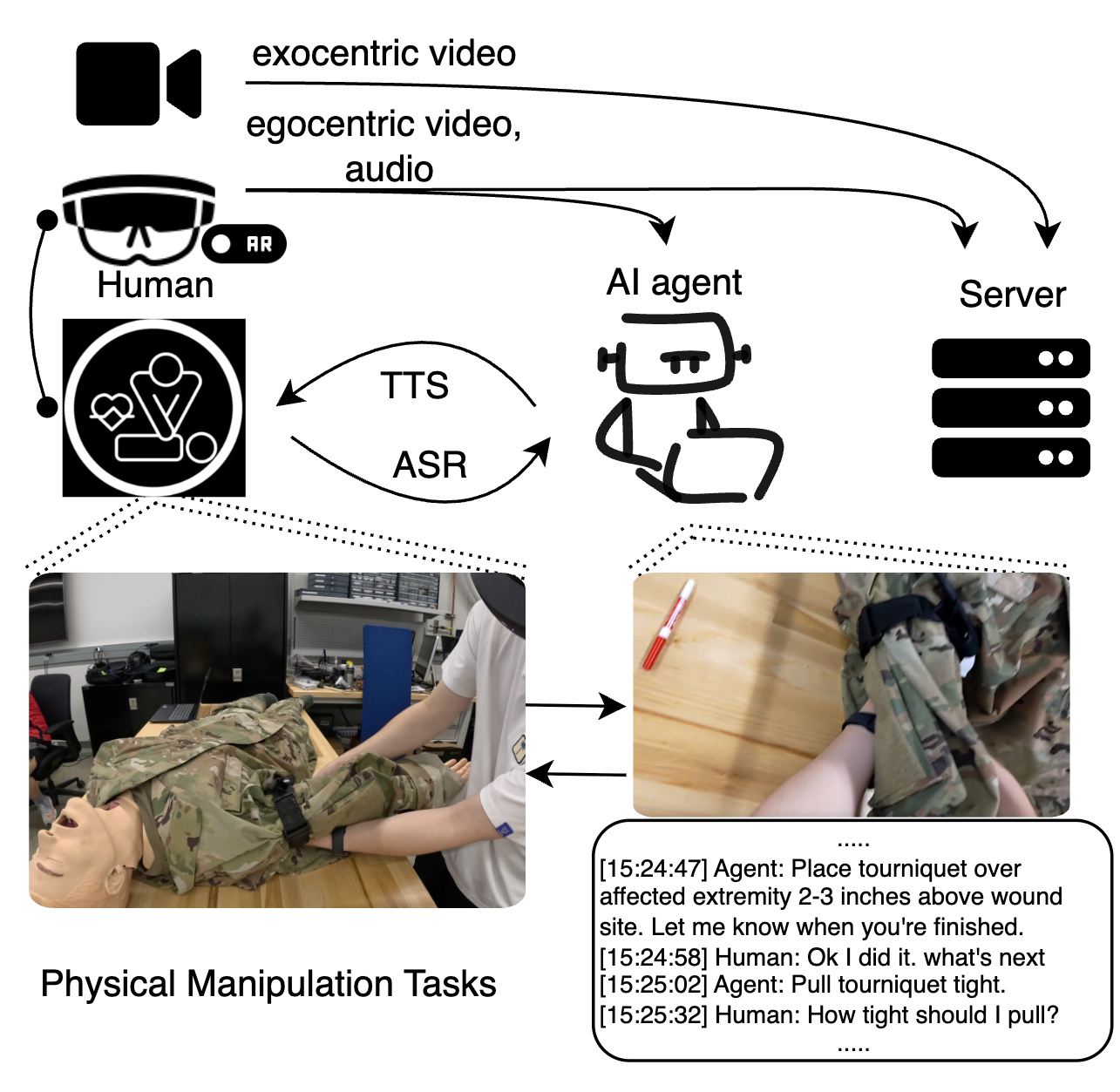} %
  \caption{
  Workflow of an interactive AI agent guiding a human through a physical task. The human performs the task (e.g., applying a tourniquet) in the physical world while communicating with the AI agent, which is delivered through an AR headset.
  }
    \label{data collection} 
  \end{figure}

Recent progress in Artificial Intelligence (AI) have been driven by the rapid development of Large Language Models (LLMs) such as GPT~\cite{openai2024gpt4technicalreport}, Claude~\cite{TheC3}, Gemini~\cite{geminiteam2025geminifamilyhighlycapable}, Qwen~\cite{Bai2023QwenTR,Yang2024Qwen25TR} and LLaMA~\cite{touvron2023llama}, which exhibit impressive capabilities in language understanding, reasoning, and complex task execution. The integration of multimodal capabilities into these models has further expanded the reach of AI, enabling systems to process and generate diverse content across text, images, and video. Despite these advancements, the real-world deployment of AI agents in human-in-the-loop scenarios—where AI collaborates with humans in completing physical tasks—remains an underexplored domain, particularly in dynamic and interactive environments.
(Fig.~\ref{data collection}).

Human-AI collaboration in assisted autonomy is emerging as a critical paradigm, with AI agents providing context-aware guidance and interactive feedback to enhance human task completion~\cite{li2024instructional,bellos2024can,storks2024explainable,huang2024position,putta2024agentqadvancedreasoning,kapoor2024ai,lu2024ai,xi2023rise}. While AI-powered agents have demonstrated effectiveness in controlled settings, their application in high-stakes, physically dynamic tasks—such as medical procedures—presents unique challenges. Augmenting human cognition through AI-equipped augmented reality (AR) agents holds promise for both professional domains, such as battlefield medicine and surgical assistance, and everyday activities, such as cooking and assembly. However, a major limitation in advancing AI-assisted collaboration is the absence of structured evaluation frameworks that rigorously assess AI agents’ effectiveness in human-in-the-loop task completion beyond traditional automation benchmarks~\cite{kapoor2024ai,wang2024mmlu,li2024genai}.

To address this gap, we introduce a comprehensive evaluation framework specifically designed to assess AI-guided physical task completion in real-world human-AI collaboration. While prior evaluation methods largely focus on language models or simulated tasks, they fail to capture the complexities of embodied assistance—such as interactive guidance, execution quality, user experience, and learning outcomes. Our framework defines key metrics for quantifying AI effectiveness across these dimensions and is grounded in a newly collected multimodal dataset of human-AI interaction sessions. The dataset includes synchronized egocentric and exocentric video, audio, and rich annotations of step-level outcomes, error categories, and natural language rationales—enabling fine-grained analysis of procedural support and user adaptation. While not a universal benchmark, this framework represents a critical first step toward rigorous and reproducible evaluation of embodied AI agents in physical environments.

To support this framework, we develop an AI-powered system for real-world physical task collaboration. At its core it is a perceptually enabled agent that integrates AR-based guidance with real-time task monitoring and feedback. This system allows us to perform and evaluate AI-guided assistance across a range of tasks, from everyday activities to high-stakes domains such as battlefield medicine.
Building on this foundation, we conduct extensive human studies to analyze how users interact with AI-guided systems in physical environments. Our studies examine adaptation to AI workflows, the cognitive impact of interactive guidance, and the design principles that support effective collaboration. These insights deepen our understanding of AI agents as collaborative partners rather than passive automation tools and highlight critical challenges and opportunities in real-world deployment.

Our contributions are threefold:

\begin{enumerate}
    \item A structured evaluation framework and dataset for measuring AI-guided physical task performance;

    \item An interactive AR-based AI agent that serves as a key reference for real-world AI task guidance;

\item Empirical findings and dataset from human studies on performance, experience, and workflow design in AI-assisted task completion.
\end{enumerate}

\section{Related Work}

\paragraph{Foundation Models and AI Agents}
Recent advances in Large Language Models (LLMs) have spurred the development of Multimodal LLMs (MLLMs) capable of handling inputs beyond text—such as images, audio, and 3D data—with notable examples including OpenAI ChatGPT~\cite{openai2024gpt4technicalreport}, LLaVA~\cite{NEURIPS2023_6dcf277e}, MiniGPT-4~\cite{zhu2024minigpt}, LLaMA-Adapter~\cite{zhang2024llamaadapter}, and Google’s Gemini~\cite{geminiteam2025geminifamilyhighlycapable}. These models extend vision-language reasoning and are promising for task guidance requiring contextual understanding. However, despite strong performance in structured or virtual settings, they often struggle in real-world human-in-the-loop scenarios that demand physical adaptability and real-time interaction. While prior work like “Watch, Talk and Guide” (WTaG)~\cite{bao-etal-2023-foundation} explores foundation models's zero-shot capability for cooking task guidance, they largely overlook the need of efficient communication in more challenging domains. In contrast, our work deploys a fully implemented AR-equipped efficient AI agent with ad-hoc ML modules for physical collaboration across domains---from cooking to battlefield medicine---establishing a foundation for assessing smoothness, robustness, and adaptability of AI in human-AI interactions for professional scenarios.

\paragraph{AI Agents Evaluation Framework}
Evaluating AI systems is essential to ensure reliable performance in real-world scenarios, especially as generative AI expands into task-guidance applications. Existing benchmarks such as Big Bench Hard (BBH)~\cite{suzgun2022challenging}, MMLU-PRO~\cite{wang2024mmlu}, and GenAI-Bench~\cite{li2024genai} offer robust evaluations for language and vision models but largely focus on virtual tasks. Similarly, datasets like IFEval~\cite{sprague2023musr} and MuSR~\cite{zhou2023instruction} test cross-modal and instruction-following capabilities, yet fail to capture the interactive and dynamic nature of real-world human-AI collaboration.

Task guidance systems require evaluation frameworks that reflect physical interaction, real-time decision-making, and user adaptation. 
While works like "Watch, Talk and Guide" (WTaG)~\cite{bao-etal-2023-foundation} collect valuable human-in-the-loop data and assess user intent understanding, they primarily focus on evaluating foundation models' capabilities rather than measuring the practical effectiveness of AI guidance systems.%
Our framework fills this gap by incorporating metrics for task success, error reduction, user satisfaction, and robustness, enabling a more practical and holistic evaluation of embodied AI agents.

\section{Evaluation Framework for Human-AI Collaborative Task Completion}
Our framework evaluates both task completion performance and the quality of user interaction in AI-assisted task guidance scenarios. It is supported by a multimodal dataset comprising synchronized egocentric and exocentric recordings, with detailed annotations from both AI-assisted and unassisted task executions across multiple tasks.

\subsection{Task Completion Quality Assessment}

This component evaluates the AI assistant's ability to provide accurate and timely guidance. We define the following metrics:

\begin{itemize}[leftmargin=*,itemsep=0pt]
    \item \textbf{Success Rate Metrics:} We measure success at two levels:
    \begin{itemize}[leftmargin=*,itemsep=0pt]
        \item {Macro Success Rate (M-SR):} This metric calculates the average success rate across all samples from all tasks combined. It provides a global view of the AI's effectiveness across the entire dataset.
        \item {Micro Success Rate ($\mu$-SR):} This metric first calculates the average success rate within each task separately, then averages these task-specific rates. This approach ensures equal weighting of tasks regardless of their sample sizes and provides insight into the AI's performance across different task types.
    \end{itemize}
    Both metrics help quantify the efficiency gains (or losses) introduced by the AI assistant, especially when compared to traditional (non-AI) methods. Higher success rates indicate that the AI is successfully helping users navigate tasks from start to finish.

    \item \textbf{Time to Completion:} We measure the average time taken to complete the task with AI assistance compared to unassisted settings. A reduction in completion time suggests that the AI is effectively streamlining the task process.

    \item \textbf{Step Error Rate (S-ER):} We track errors from the user's perspective, both when guided by AI and when not. We categorize errors into two main types:
    \begin{itemize}[leftmargin=*,itemsep=0pt]
        \item{Critical Errors:} These are errors that render the completion of the task impossible. They represent significant deviations from the correct procedure that cannot be recovered from without starting over.
        \item{Step-Specific Errors:} These are more fine-grained errors that occur within individual steps of the task. These errors could be Wrong Action (e.g., stirring instead of folding), Wrong Object (e.g., using salt instead of sugar), Wrong State (e.g., over-beating eggs) or Other step-specific errors that don't fall into the above categories.
    \end{itemize}

    \item \textbf{Error Reduction:} We report the Step Error Rate (S-ER) for both AI-assisted and unassisted task completion conditions. These measurements allow us to assess the AI's impact on error reduction across different guidance methods.

    \item \textbf{Step-Guidance Alignment:} This measures the rate at which the AI provides instructions that correctly correspond to the current step of the task. High alignment indicates that the AI is maintaining awareness of the user's progress and providing contextually appropriate guidance.
\end{itemize}
\begin{figure*}[t]
  \centering
  \hspace*{-0.7cm} %
  \includegraphics[width=0.9\textwidth]{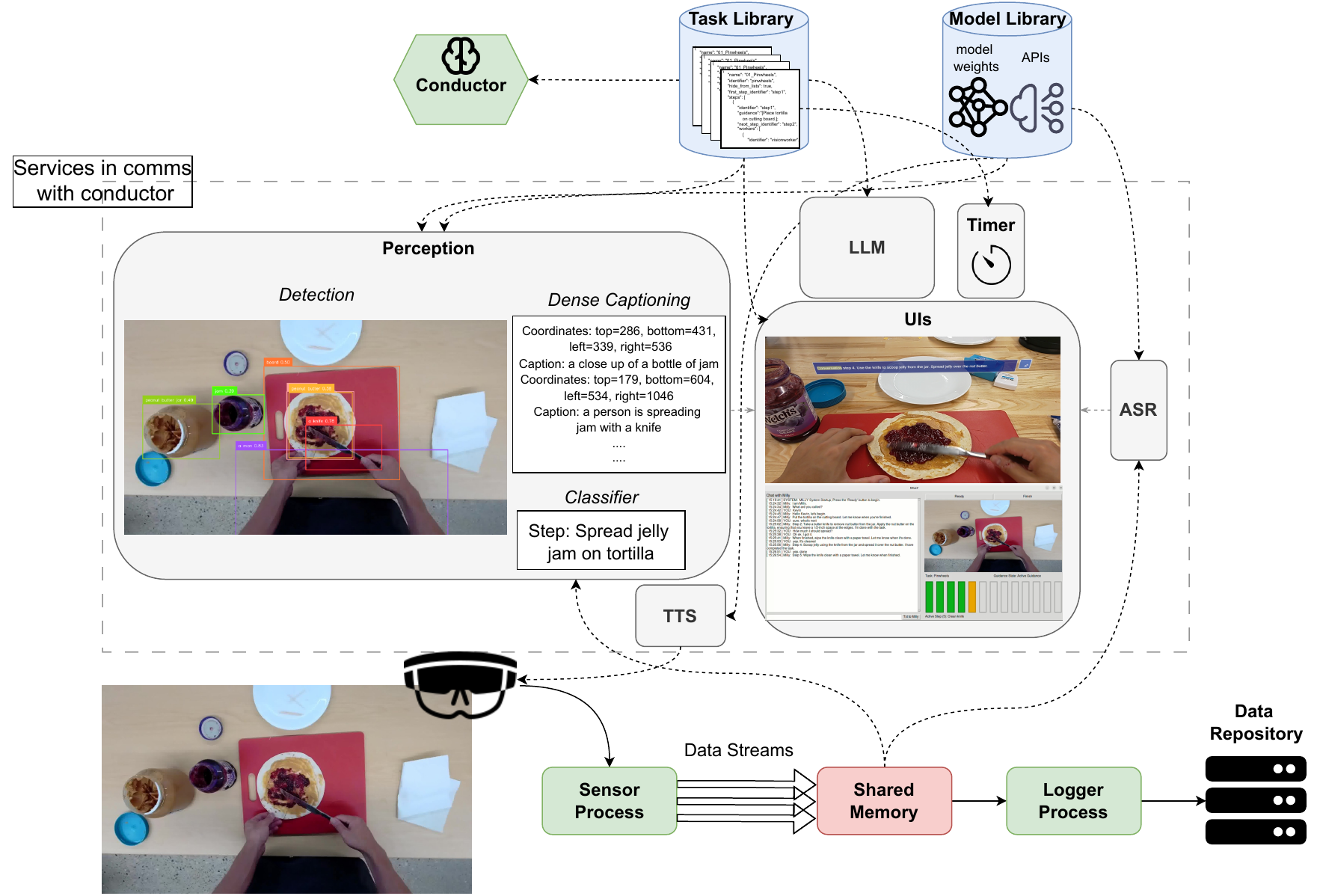} %
  \caption{
Overview of the Baseline Interactive Agent for Physical Task Assistance.
}
  \label{system}
\end{figure*}

\subsection{User Interaction Quality Evaluation}

User experience is paramount to the success of AI agents, and it will eventually determine their effectiveness and adoption. Evaluating user interaction metrics is essential for understanding how users engage with the agent, as well as for identifying areas where the system can be improved to better meet user needs and expectations.

\begin{itemize}[leftmargin=*,itemsep=0pt]
    \item \textbf{Clarity:} Users rate the clarity and comprehensibility of AI-generated instructions. This metric is crucial for understanding how well the AI communicates its guidance and whether users can easily interpret the provided instructions.

    \item \textbf{Proactivity:} We evaluate the AI's ability to offer timely, unsolicited guidance. Users assess these interventions based on their helpfulness and appropriateness to the current task context.

    \item \textbf{Ease of Use:} Users evaluate the intuitiveness and user-friendliness of the AI interface. This metric helps assess the accessibility of the AI assistant to users of varying technical expertise.

    \item \textbf{Satisfaction:} This holistic metric captures the user's general impression of the AI's performance and usefulness.

    \item \textbf{Relevance:} We measure this through query-response appropriateness - how well the AI provides relevant answers to user questions and maintains context-appropriate guidance.

    \item \textbf{Overall Score:} A comprehensive metric combining all aspects of the user interaction experience.
\end{itemize}

\subsection{Cost-Controlled Performance Evaluation}
\label{sec:cost}
Balancing AI performance with computational and financial costs is crucial for scalable deployment, particularly in human-in-the-loop and constrained environments such as battlefield medicine and home assistance. High-performing AI agents, especially those leveraging Large Language Models (LLMs), can incur significant inference costs that may not always translate to proportional benefits~\cite{kapoor2024aiagentsmatter}. To address this, we introduce a \textbf{Cost vs Performance} evaluation framework, incorporating \textbf{Inference Cost}, which quantifies computational and monetary expenses, and \textbf{Cost-Performance Pareto Efficiency}, which identifies optimal trade-offs between resource consumption and guidance effectiveness.

\subsection{Evaluation Methodology}

Our evaluation process involves collecting video recordings of tasks, AI-user conversation logs, and post-task user surveys.

Evaluation points are systematically extracted from various stages of the interaction, including user inputs, AI responses, and task execution events. This approach allows us to capture the dynamic nature of the interaction, assess the AI's performance in various scenarios, and identify potential blind spots in both the AI's guidance and the user's awareness. By including unnoticed errors as evaluation points, we can assess the AI's ability to detect and prevent mistakes, as well as the user's reliance on the system for error prevention.

The functional metrics are calculated based on objective measures from the recordings and logs, while the user interaction metrics combine objective measures with subjective user ratings from the surveys. This comprehensive data collection and analysis strategy ensures a thorough evaluation of the AI assistant's performance across different aspects of task guidance.

By providing clear definitions and measurement methods for each metric, we ensure that this framework can be consistently applied and replicated in future studies, facilitating comparative analysis across different AI assistant implementations. This standardized approach will contribute to the ongoing improvement of AI-assisted task guidance systems, ultimately leading to more effective and user-friendly solutions.

\section{Human-AI Collaborative Agent}
\label{sec:system}
We explore the application of pre-trained Large Language and Vision Foundation Models on this problem without task-specific training. We propose an AI agent architecture and three different configurations to extract visual and dialog context:

\subsection{System Design and Development}

Our task guidance agent (Figure~\ref{system}) is built upon a modular, multi-process architecture designed for robustness, scalability, and real-time performance. At the core of our agent is the \textbf{Conductor Process}, which orchestrates the overall workflow. All major components—including Perception, LLM, UIs, TTS, ASR, and Timer—communicate directly with the Conductor but also with each other through it.
It listens to processed input information from the environment and user, performing node transitions based on predefined conditions. This process effectively acts as a state machine, managing the flow of the task guidance and ensuring seamless task progression and coordination across all system components.

At the beginning of each new user session, the Conductor accesses the \textbf{Task Library} to build a task graph based on the specific task the user has indicated they plan on performing. This task graph, constructed from the information stored in the Task Library, serves as the backbone for guiding the user through the process.

Working in tandem with the Conductor is the \textbf{Data Manager Process}. This component, represented as the \textbf{Sensor Process} and \textbf{Shared Memory} in our architecture, continuously pulls data from sensors and peripherals, storing it in shared memory and making it readily accessible to other processes. By centralizing data handling, we ensure that all components of the system have access to the most up-to-date information, crucial for maintaining coherence in task guidance.
Output is managed by the \textbf{UIs (User Interfaces)}, with the \textbf{HL2 Output Process} rendering information to the Hololens 2,providing an immersive augmented reality experience for the user, and the \textbf{Naive Output Process} displaying it on a separate monitor(debugging, monitoring).

Visual understanding is handled by the \textbf{Active Perception Process}, which continuously processes visual data based on the current node's information. This adaptive approach allows the system to focus its computational resources on the most relevant aspects of the visual input, enhancing efficiency and accuracy in scene interpretation. Importantly, the Perception Service can detect if a user performs a step out of sequence, triggering an alert to the Conductor, which then prompts the AI agent to enter a conversation mode. In this mode, the AI interacts with the user to assess the situation and determine if there is an issue that needs to be addressed. We specifically conduct the evaluation on this Process in Section~\ref{sec:perception}.

Natural language processing is a key component of our system, managed by the \textbf{LLM Process}. This process provides critical services such as categorizing free-form user input and rephrasing canonical information into natural language. By leveraging advanced language models, we enable intuitive interaction between the user and the system.

Audio input and output are handled by two specialized processes. The \textbf{ASR Process} constantly processes audio from the microphone, broadcasting processed text when necessary. This allows for hands-free interaction, crucial in many task guidance scenarios. Complementing this, the \textbf{TTS Process} converts text to audio upon request from the Conductor process, pushing it to the Output process for vocalization. This enables the system to provide auditory guidance, enhancing the multi-modal nature of the interaction.

The \textbf{Timer Service} monitors task duration and can interrupt the user if a predefined time threshold is exceeded. When triggered, the Timer Service prompts the AI agent to enter conversation mode, engaging the user to check if anything is wrong or if assistance is needed. This feature ensures safety and efficiency in time-sensitive tasks or when prolonged inactivity might indicate a problem.

A \textbf{Logger Process} continuously accesses the Shared Memory, recording all relevant data streams to the \textbf{Data Repository}. This comprehensive logging facilitates system analysis, performance optimization, and continuous improvement of the task guidance algorithms.

The architecture leverages shared memory and ZeroMQ for inter-process communication, optimizing both high-throughput data sharing and low-latency message passing. This hybrid approach ensures efficient data transfer and system responsiveness, which are critical for real-time task guidance.

\section{Dataset Collection}
\label{sec:dataset-collection}

We captured synchronized multi-modal egocentric-exocentric view recordings for 4 tasks by 12 participants. In addition to step time boundaries, we provide mistake detection annotations at task and step level with natural language descriptions. These media and annotations are not only useful for human action understanding training but also human-AI collaboration.

\begin{figure}[t] 
    \centering
    \hspace*{-0.7cm} %
    \includegraphics[width=0.45\textwidth]{
    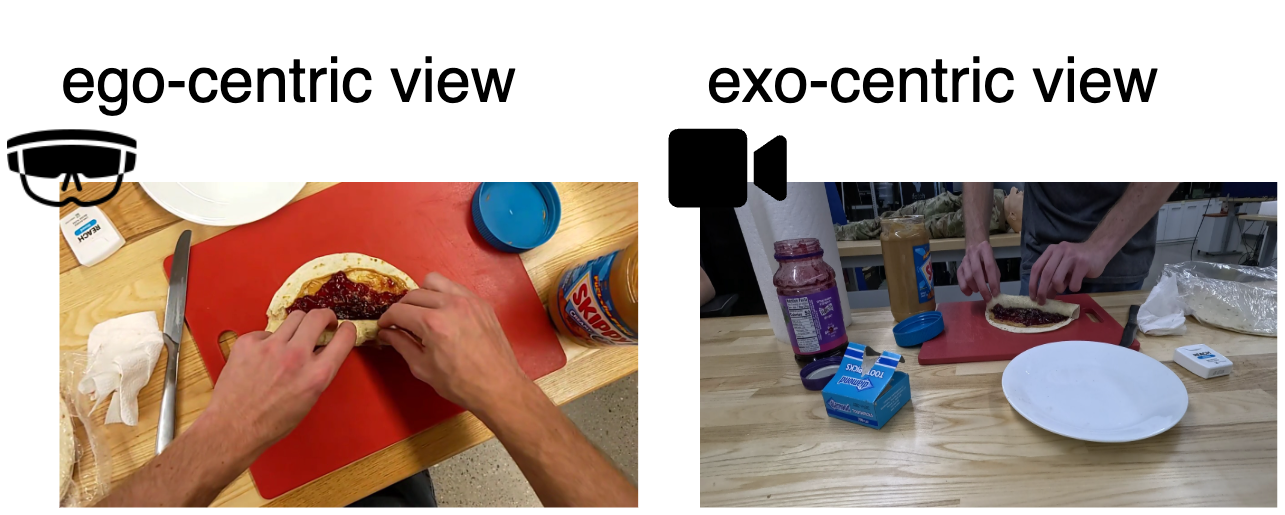
    } %
  \caption{
  Example from our dataset. Left: Microsoft Hololens 2 view. Right: GoPro Hero 12 view.
  }
    \label{fig:data_collection} 
  \end{figure}

\textbf{Participants} 
We recruited 12 participants (6 male, 6 female; ages 19–29) from a U.S.-based university. The cohort was ethnically and culturally diverse: 5 participants identified as Chinese (including 3 international students and 2 U.S.-raised), 3 as Indian/South Asian, 3 as white American, and 1 as mixed Chinese–European descent. Academic standing ranged from first-year undergraduate to doctoral level, with 7 undergraduates and 5 graduate students (3 M.S. and 1 Ph.D.) whose majors spanned robotics, computer and electrical engineering, aerospace engineering, neuroscience, and kinesiology, with several pursuing dual concentrations. All participants reported normal or corrected-to-normal vision. This breadth of cultural background and disciplinary expertise provided a rich and varied pool for data collection.

\textbf{Tasks} Each participant completed four tasks that span everyday cooking to battlefield medicine: \emph{make a cup of tea}, \emph{prepare pinwheel sandwiches}, \emph{prepare a dessert quesadilla}, and \emph{apply a tourniquet}.
For every task we evaluated three assistance conditions.
In the \textbf{Unassisted} participants were provided only with a task name and a brief goal description, relying solely on prior knowledge.
In the \textbf{Paper-Instructions} condition users were given a detailed, step-by-step description of the task. This method represents traditional cookbook-style instructions (\cref{tab:paperass}).
In the \textbf{AI-Agent} condition, participants receive interactive, context-aware guidance via our AR-based AI system (\cref{sec:system}).

\textbf{Recordings}
There 144 sessions in total where one session records one participant performing one task with one type of instruction. For each session, we recorded a 3-rd person view video of the participant's actions with GoPro Hero 12 Black camera. For sessions AI instruction, we also recorded the participant's first-person view and dialogue with a Microsoft Hololens 2 with data from all sensors, including the front camera, 4 side cameras, depth sensor, IMU, and audio. 
The average duration of sessions are 4.73$\pm$2.01 minutes, resulting in a total of 15.15 hours of valid experiment duration (the duration of exo-centric view recordings of all sessions). We record all sessions in one room. An illustration of egocentric-exocentric view (front RGB camera) of ``Pinwheel'' task is illustrated in~\cref{fig:data_collection}.

\textbf{Annotations}
Two trained annotators with undergraduate-level engineering backgrounds labeled the video recordings, following a predefined protocol. The annotation process was conducted using the VGG Image Annotator (VIA) tool~\cite{dutta2016via, dutta2019vgg}, which allows for efficient video annotation and supports temporal segmentation. Our annotations capture \emph{five} complementary data fields, each stored in a machine-readable JSON file to facilitate downstream analysis:
(1)~\textbf{Task success (Boolean)} and a free-form \texttt{comment} string when it is a failure.
(2)~\textbf{Task duration} is recorded as two floating-point timestamps (\texttt{start\_sec}, \texttt{end\_sec}) that mark the interval from the participant’s initial engagement with the instructions to their explicit confirmation of task completion.
(3)~\textbf{Step boundaries} are entries where each contains \texttt{start\_sec} and an \texttt{end\_sec}. A new step begins when there is unmistakable visual evidence of its tacit action (e.g., the hand moves toward a paper towel to initiate “clean knife”).
(4)~\textbf{Out-of-order mistake} is flagged by comparing the temporal order of the annotated steps against the canonical recipe order.
(5)~\textbf{Fine-grained mistakes inside a step}—such as using the wrong tool or mismeasuring an ingredient—are logged as a list of \texttt{\{"step \#": "free-form description"\}} objects.
(6)~\textbf{Synchronization}---for sessions with egocentric and exocentric recordings, we manually synchronize the two streams by annotating the time offset, ensuring accurate temporal alignment across views.

\section{Human Study on AI-Assisted Task Collaboration}

We conducted a structured user study to evaluate how different guidance methods influence physical task performance, learning outcomes, and user experience. Participants completed real-world tasks using the same four scenarios and guidance conditions introduced in Section~\ref{sec:dataset-collection}.

\subsection{Experimental Design}

Each participant completed the same task three times, once under each of the three guidance conditions: Unassisted (UA), Paper Instructions (PI), and AI Agent (AI). To control for ordering effects, we employed full counterbalancing across the six possible permutations (e.g., UA $\rightarrow$ PI $\rightarrow$ AI, UA $\rightarrow$ AI $\rightarrow$ PI, etc.), with participants randomly assigned to one of the orders.

Tasks were selected from those used in dataset collection: three recipe-style procedures and one medical scenario. We strategically chose tasks to span a range of complexity levels. The recipe tasks—\emph{make tea}, \emph{pinwheels}, and \emph{dessert quesadilla}—are commonly used to evaluate instructional systems due to their clear temporal structure and objective success criteria. To validate our system’s applicability in higher-stakes domains, we included \emph{tourniquet application}, a significantly more complex and safety-critical task from battlefield medicine.

\textbf{Participants.}
This study involved 12 participants (distinct from annotators), each completing multiple task variants under different guidance conditions.

\textbf{Exposure Consideration.}
Since each participant performed the same task multiple times under different guidance methods, performance may be influenced not only by the guidance condition itself but also by prior exposure to the task. To account for this, our analysis considers both first-time and repeated attempts, allowing us to examine learning effects and how prior experience with one method impacts performance under another.

\subsection{Evaluation Results}
We report the performance of different methods of task guidance in terms of functional performance, user interaction quality, and skill acquisition.
We present the results for physical task completion according to our proposed evaluation framework. Results are interpreted using our proposed evaluation framework and analyzed with respect to participants' task exposure.

\textbf{Task Completion Quality Assessment.} The task performance assessment results, presented in Table \ref{tab:task1}, clearly demonstrate the effectiveness of AI-assisted guidance. 

When users attempted tasks the first time without any prior training (Training=None), those guided by the AI system achieved a significantly higher Macro Success Rate (M-SR) of 70\%, compared to only 20\% with unassisted guidance (UA) and 28.57\% with paper instructions (PI). The Step Error Rate (S-ER) followed a similar trend, favoring AI at 16.43\%, over PI (18.37\%) and UA (38.75\%). While AI-guided tasks took longer to complete (186.54 seconds), the tradeoff favored higher success and lower error.

We also examined how initial exposure to AI guidance influenced subsequent performance using other methods.

\begin{table}[h]
    \centering
    \begin{tabular}{cccccc}
        \toprule
        \textbf{Training} & \textbf{Guidance} & \textbf{M-SR} $\uparrow$  & \textbf{S-ER} $\downarrow$ & \textbf{Time(s)} $\downarrow$ \\
        \midrule
        \multirow{3}{*}{\textbf{None}}
            & {UA} & {20.00\%} & {38.75\%} & \textbf{106.26}  \\
            & {PI} & {28.57\%} & {18.37\%} & {144.29}   \\
            & {AI} & \textbf{70.00\%} & \textbf{16.43\%} & {186.54}  \\
        \midrule
        \multirow{2}{*}{\textbf{AI}}
            & {UA} & {66.67\%} & {18.45\%} & \textbf{104.77}  \\
            & {PI} & \textbf{75.00\%} & \textbf{6.70\%} & {132.29}   \\
        \midrule
        \multirow{2}{*}{\textbf{PI}}
            & {UA} & {50.00\%} & {20.09\%} & \textbf{99.80}  \\
            & {AI} & \textbf{80.00\%} & \textbf{5.00\%} & {186.69}  \\
        \midrule
        \multirow{2}{*}{\textbf{UA}}
            & {AI} & \textbf{100.00\%} & \textbf{0.00\%} & {217.49}  \\
            & {PI} & {60.00\%} & {25.00\%} & \textbf{97.82}  \\
        \bottomrule
    \end{tabular}
    \caption{Task Performance Assessment results. We present Macro Success Rate (M-SR), Step Error Rate (S-ER), and task duration (Time).}
    \label{tab:task1}
\end{table}

\begin{figure}
    \centering
    \includegraphics[width=0.82\linewidth]{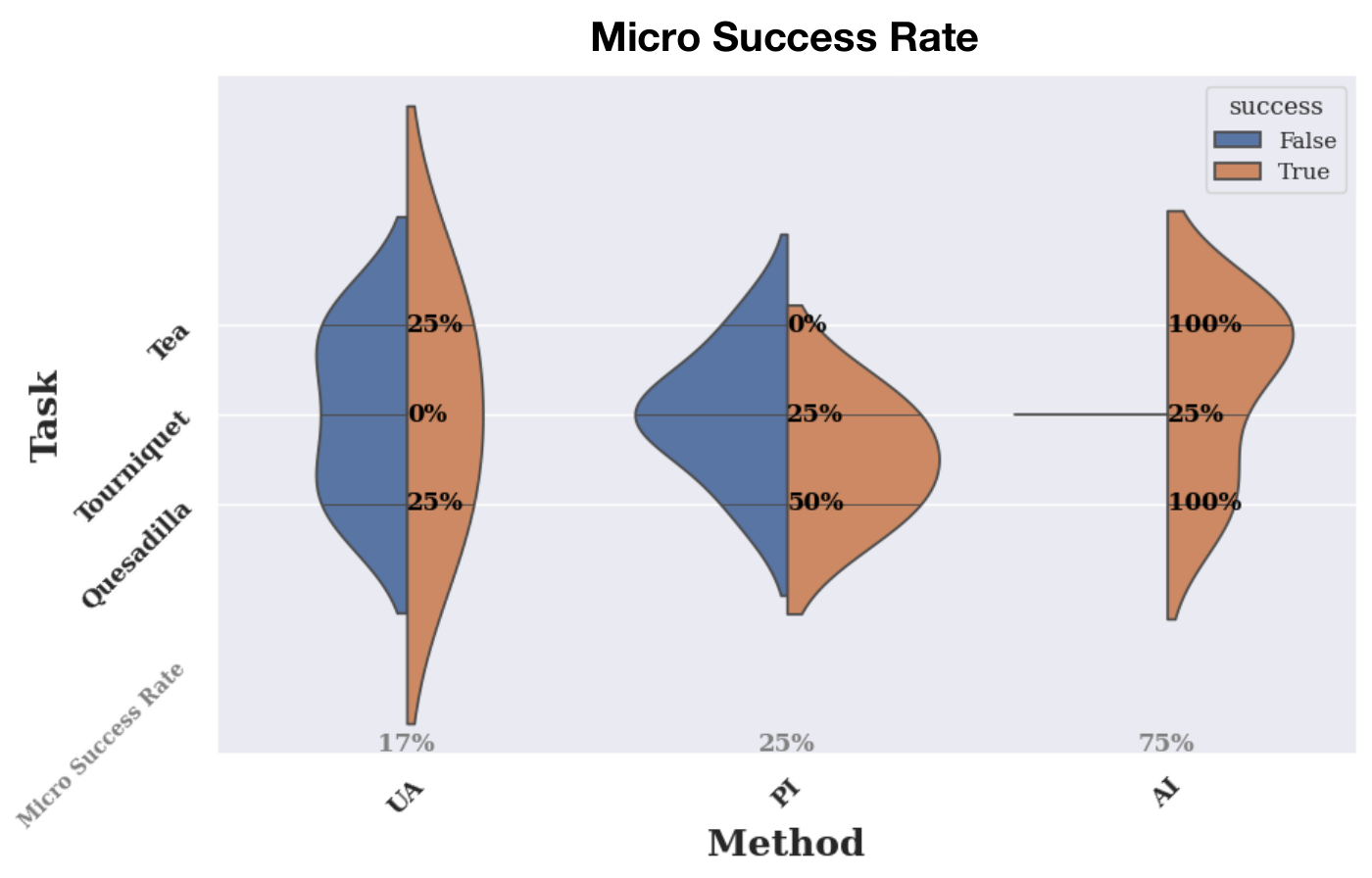}
    \caption{Micro Task Performance Assessment results. For all tasks, AI guidance help user's achieve better performance even for challenging professional tasks like Applying Tourniquet in battle field medicine. }
    \label{fig:enter-label}
\end{figure}

Specifically, Table \ref{tab:task1} presents data on \textbf{skill acquisition} across different guidance methods.

The key findings are in the "AI" condition, where participants used AI assistance in their first trial:

\begin{itemize}[leftmargin=*,itemsep=0pt]
    \item Subsequent unassisted (UA) performance dramatically improved to 66.67\% success rate, with errors reduced to 18.45\%.
    \item With paper instructions (PI) after AI exposure, performance further increased to 75\% success and only 6.70\% errors.
\end{itemize}

It is plausible that task repetition alone improves performance, as participants become more familiar with task structure and physical actions. However, we observe that improvements following AI exposure are notably greater than those following UA or PI.
The interactive and context-aware support provided by the AI system appears to facilitate more effective learning, enabling users to perform better in subsequent tasks regardless of the guidance method used. This enhanced skill acquisition underscores the potential of AI agents not just as task assistants, but as effective training tools for physical manipulation tasks.

The \textbf{Micro Task Performance} results shown in Figure \ref{fig:enter-label} reinforces this trend across individual tasks, including the most complex (tourniquet application), where AI consistently enabled better performance regardless of task domain.

\textbf{User Interaction Quality Evaluation.} We report user perceptions regarding the helpfulness of starting with the AI assistant method in completing subsequent tasks using other methods. The majority of participants, 77.8\%, reported finding the AI assistant method to be more helpful when transitioning to other methods. This suggests that the AI assistant may have provided users with a strong foundational understanding or approach that benefitted them in subsequent tasks. On the other hand, 22.2\% of participants felt that starting with the AI assistant method didn't make a difference in their ability to complete the other methods, indicating that, for some users, the AI's influence was neutral. Notably, no participants found the AI assistant method to be less helpful.

These findings indicate that the AI assistant method generally provided a positive learning experience or offered insights that users could apply to other task completion methods. The high percentage of users finding it more helpful suggests that the AI assistant may be effective in training or familiarizing users with tasks, improving their performance even when the AI is not directly guiding them, claim that is substantiated in the quantitative analysis for skill acquisition.

\begin{figure*}[t]
    \centering
    \hspace*{-0.7cm} %
    \includegraphics[width=1\textwidth]{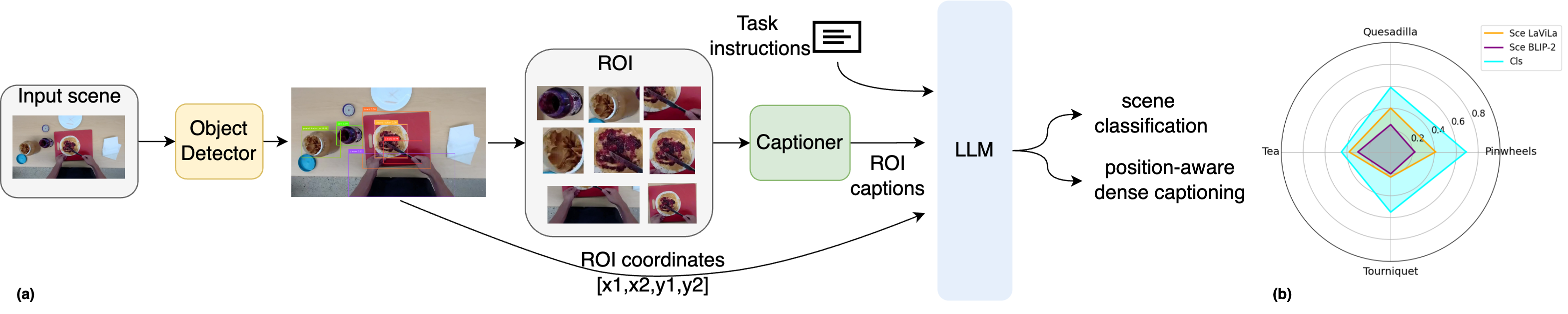}\vspace{-3mm} %
    \caption{
  \textbf{(a)}: Zero-shot scene classification and dense captioning. \textbf{(b)}: Accuracy report for perception methods.
  }
    \label{exp2}
  \end{figure*}

\begin{table}[h]
    \centering
    \begin{tabular}{ccc}
        \hline
        Question & Score Logit (\/5)  $\uparrow$  & Score Percentage  $\uparrow$ \\
        \hline
        {Clarity} & {3.42} & {68.33\%} \\
        \hline
        {Proactivity} & {3.17} & {63.33\%} \\
        \hline
        {Ease of use} & {3.08} & {61.67\%} \\
        \hline
        {Satisfaction} & {3.00} & {60.00\%} \\
        \hline
        {Relevance} & {2.67} & {53.33\%} \\
        \hline
        \textbf{Overall} & \textbf{3.07} & \textbf{61.33\%} \\
        \hline
    \end{tabular}
    \caption{User Interaction Quality Metrics.}
    \label{tab:placeholder}
\end{table}

The user interaction quality metrics presented in Table \ref{tab:placeholder} provide insights into various aspects of the AI assistant's performance as perceived by users. The metrics cover clarity, proactivity, ease of use, satisfaction, and relevance, each rated on a 5-point scale. The results indicate that the AI assistant performed best in providing clear instructions, with a score of 3.42 out of 5 (68.33\%). Proactivity in guiding users through tasks was also well-received, scoring 3.17 (63.33\%). The ease of use and overall satisfaction were rated slightly above average, both scoring around 3 out of 5 (61.67\% and 60.00\% respectively). The relevance of the AI's responses to user queries received the lowest score of 2.67 (53.33\%), suggesting an area for potential improvement. The overall performance, calculated as the mean score across all questions, was 3.07 (61.33\%), indicating that while the AI assistant generally met user expectations, there is room for enhancement.

\textbf{Cost.} Our AI task guidance system balances cost and performance efficiently, operating on a Lenovo ThinkPad P16 Gen 2 with an NVIDIA RTX 5000 GPU and a Microsoft HoloLens 2, totaling \$9,019. Each task session incurs an average inference cost of \$0.002 using OpenAI's ChatGPT~\cite{ouyang2022training}, achieving a inference cost-to-success rate ratio~\cite{kapoor2024ai} of 0.000029 \$/\%, demonstrating strong cost-effectiveness in AI-assisted task completion.

\subsection{Perception}
\label{sec:perception}
To fully understand an AI agent's potential in task guidance, it is crucial to evaluate its perception abilities separately. This isolation allows us to independently assess how well it can interpret and respond to visual inputs from the environment, a critical factor in real-time guidance. To that end, we integrated two methods in our system, one adopting a zero-shot approach, and the other a supervised.

\subsubsection{Scene Description (Sce)}

This method enhances situational awareness by generating a free-text scene description using a combination of an object detector, a captioning module, and a Large Language Model (LLM).

As we can see in Figure~\ref{exp2} (a), we apply an object detector, here we use DINO~\cite{li2022dn}, to identify regions of interest (ROI) in the latest captured frame. These ROI are then fed into a captioner, we experiment with BLIP-2~\cite{10.5555/3618408.3619222} and LaViLa~\cite{zhao2023lavila}, along with prompts to generate descriptions for each region. LaViLa which, unlike BLIP-2, takes a sequence of frames as input, allowing it to better capture temporal dynamics of the task completion.

The resulting descriptions from either captioner, along with their corresponding region coordinates, are provided as a prompt to a Large Language Model (LLM) (GPT-3.5-turbo), together with a full list of the task steps. The LLM then maps the individual descriptions to a holistic scene state recognition acting as a task step classifier but also providing a comprehensive scene description with positional information.

This zero-shot approach allows the system to generalize to diverse or unfamiliar environments where task-specific data may not be available. The classification accuracy from this approach is limited by the constraints imposed by the different components in the pipeline. For instance, the object detection and captioning stages, while effective at generating rich and detailed scene descriptions, introduce noise and variability that can affect the LLM's ability to accurately classify task steps.

\subsubsection{ResNet Step Classifier (Cls)}

To track task progress more accurately, we implement a ResNet-based step classifier. Although it requires task-specific training data, it is particularly suited for scenarios where computational resources are limited, such as in remote environments, making it valuable for task-guidance applications in specialized domains like battlefield medicine.

As expected, Cls outperforms Sce across tasks, owing to its task-specific training and the in-distribution test environment data(Figure~\ref{exp2} (b)). However, Sce accurately detects salient regions (not quantitatively evaluated here), which could prove highly valuable in future iterations of our system. Additionally, fine-tuning the captioner is likely to yield significant performance improvements, which we plan to explore in future work.

\section{Conclusion}
	For human-AI collaborative task completion, we introduced a comprehensive evaluation framework and developed an AR-equipped AI agent for interactive guidance. Our human studies validate the framework’s effectiveness, providing valuable insights into AI-assisted collaboration from both a technical design and human learning perspective across diverse task scenarios. First, the results demonstrate that our AI agent significantly improves task success and reduces errors. Additionally, our studies offer deeper insights into human skill development in AI-assisted settings, revealing how AI guidance shapes learning curves, task adaptation, and user confidence. These findings underscore AI’s potential not only for improving task performance but also for facilitating structured skill acquisition. Furthermore, we contribute to the research community by sharing anonymized multimodal data from our human study, along with expert-labeled task assessments, enabling further analysis and benchmarking. Based on our findings, future work can explore intuitive human query interfaces, advanced perception models, and proactive intervention strategies to enhance adaptability and user experience in human-AI collaboration for task completion.

\newpage

\bibliographystyle{ieeenat_fullname}
\bibliography{main}

\newpage

\end{document}